\documentclass[runningheads]{llncs}

\usepackage{eccv}

\usepackage{eccvabbrv}

\usepackage{graphicx}
\usepackage{booktabs}

\usepackage[accsupp]{axessibility}  
 
\usepackage{pifont}     
\usepackage{makecell}
\usepackage{array}

\usepackage{hyperref}
\usepackage{orcidlink}

\newcommand{\myparagraph}[1]{\vspace{0pt}\noindent{\bf #1:}}

\begin{document}

\title{3D Scene-Adaptive Trajectory-Controllable Human Image Animation with Camera Movement} 

\titlerunning{3STC-HIA}

\author{Deyin Liu\inst{1,4}\orcidlink{0000-0002-0371-9921} \and
Jicheng Xu\inst{1}\orcidlink{0009-0008-1929-5482} \and
Lin Yuanbo Wu\inst{2,3}\orcidlink{0000-0001-6119-058X} \and
Xiaowei Zhao\inst{2}\orcidlink{0000-0002-1182-4502} \and
Xiatian Zhu\inst{4}\orcidlink{0000-0002-9284-2955} \and
Zhe Jin \inst{1,}\thanks{Corresponding author.}\orcidlink{0000-0003-4501-7992}
\and
Anjan Dutta\inst{4}\orcidlink{0000-0002-1667-2245}}

\authorrunning{D.~Liu et al.}


\institute{
Engineering Research Center of Autonomous Unmanned System Technology, Ministry of Education; Anhui Provincial Key Laboratory of Security Artificial Intelligence; and School of Artificial Intelligence, Anhui University, Hefei, China\\
\email{iedyzzu@outlook.com}, \email{wa24301183@stu.ahu.edu.cn}, \email{jinzhe@ahu.edu.cn} \and
University of Warwick, Coventry, United Kingdom\\
\email{\{yuanbo.lin, Xiaowei.Zhao\}@warwick.ac.uk} \and 
Zhejiang Yuexiu University, Shaoxing, China \and
University of Surrey, Guildford, United Kingdom\\
\email{\{xiatian.zhu, anjan.dutta\}@surrey.ac.uk}
}

\maketitle
\setcounter{footnote}{0}

\begin{abstract}

Human image animation, which aims to generate a video of a reference subject following a provided action sequence, has received increasing research interest. With the development of diffusion-based/flow-based video foundation models, existing animation works have began to upgrade the guidance information from 2D skeleton/pose to 3D modeling conditions. Despite achieving reasonable results, these approaches face challenges in synthesizing trajectory-controllable human motion within natural scene under changed camera views. In this work, we present a scene-adaptive human image animation framework that controls both human motion and camera trajectories within a reconstructed 3D environment for video generation. To achieve this, we first develop a ground-adaptive 3D motion retargeting approach to enable user-friendly motion trajectory control adapting to the changes of elevations of ground and orientations automatically. Then we design a viewpoint-adaptive latent fusion mechanism to inject point-cloud geometric priors through scene-visibility masking into the generative process, providing precise guidance of viewpoint changes under camera control. Experiments on two standard human image animation benchmark datasets demonstrate remarkable improvements of our method over the state of the arts in related video generation metics. Project page: \href{https://robinhood256100.github.io/web-disp}{https://robinhood256100.github.io/web-disp}

\end{abstract}

\begin{figure*}[t]
  \centering 
  \includegraphics[width=\linewidth]{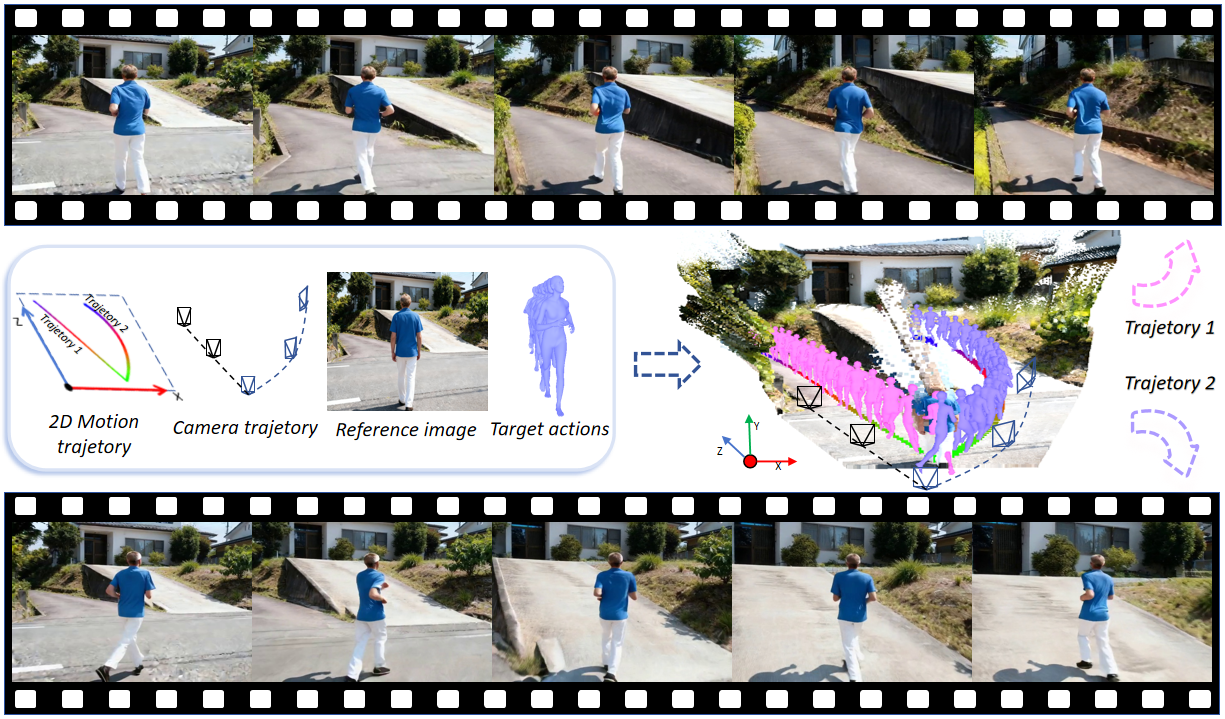} 
  \caption{\textbf{3STC-HIA} is an inference-time guidance-based human image animation method that generates scene-adaptive, trajectory-controllable human motion videos with camera movements. Given a reference image, target actions, a human motion trajectory, and a camera trajectory, the method enables human motion retargeting with ground-adaptive trajectory control, while enhancing effect of camera movement through point-cloud-based scene visibility analysis. In the top row, a subject is shown running on flat terrain along motion trajectory 1, captured from varying viewpoints defined by the camera trajectory. In contrast, the bottom row illustrates the same running actions performed on a sloped surface along motion trajectory 2 under the same camera views, where the character automatically adjusts its altitude and body orientation to adapt to the scene geometry.} 
  \label{fig:method1} 
\end{figure*}

\section{Introduction}
\label{sec:intro}

Human image animation (HIA) \cite{wang2025unianimate,xu2024magicanimate} has attracted extensive research interest due to its broad applications in social media, film production, and entertainment \cite{lin2025omni}. 
Early diffusion-based methods typically adopt 2D skeletons \cite{Yang2023Effective} as the primary motion condition for human animation video generation. These motion cues are encoded through pose-conditioning networks \cite{ZhangRA23} and injected into diffusion backbones to enforce per-frame pose and identity consistency \cite{musepose,hu2023animateanyone,zuo2024edityourmotionspacetimediffusiondecoupling,zhu2024champ,ma2024followposeposeguidedtexttovideo,wang2023disco}. 
However, only 2D pose guidance without depth information and world-space modeling, cannot effectively handle human motion and camera movement in real-world scenarios, often producing skeleton–appearance conflicts and artifacts for complex motions under viewpoint changes \cite{tu2024motionfollowereditingvideomotion,tan2024animatexuniversalcharacterimage}.

Recently, \cite{zhu2024champ,xu2024magicanimate,niu2025anicrafter} employ 3D human representations such as SMPL \cite{loper2015smpl} to explicitly model body structure with depth information, achieving more physically coherent action control.
And \cite{zhou2025realisdance-dit,liang2025realismotion,cao2025uni3c} further attempt to fit human motion into a unified 3D coordinate system, which explicitly models scene geometry, to overcome the depth ambiguity and view inconsistency of 2D-based methods. All these efforts lay foundation for the independent controls of human motion and camera movement in world space. 
However, existing methods still suffer from three challenges: (1) enabling user-friendly arbitrary human motion trajectory control which is decoupled from the source actions \footnote{To make a distinction, in this paper, we use "Action" to describe the posture and "Motion" to represent the overall human movement including the posture, motion trajectory and direction, etc.}; (2) enhancing the adaptivity to 3D scene geometry to avoid human-scene conflicts or disharmony such as floating above or penetration into the ground; and (3) considering camera dynamics and human motion simultaneously to  address the inter-frame flicker and maintain physical consistency. 

In this work, we propose a highly controllable Human Image Animation framework that achieves both 3D Scene-adaptive human motion Trajectory control and Camera movement for video generation, dubbed as 3STC-HIA. We first align a human action sequence with the reconstructed point cloud of a reference subject-scene image in the world-coordinate space. Being not content with just animating the subject conforming to the sequential actions, we also desire to control the human motion trajectory by a user-drawn plane curve. Note that simply mapping the 2D motion trajectory into the 3D world-coordinate system to operate the actions easily leads to height disharmony such as human floating or penetration when meeting with uneven ground of scene as shown in \cref{fig:height}. So exploiting elevation estimate and spatial geometry, we propose a novel human motion retargeting approach to enable ground-adaptive 3D motion trajectory control, as shown in \cref{fig:method1}. 

We further consider a more complex situation that the effect of camera movement is generated simultaneously. We observe that under changed camera viewpoints, the viewpoint-adaptive visible scene point cloud could be a useful clue to inject the real scene information in the generated video of view variations. To this end, given the 2D conditions (rendered from the aligned 3D motions and point cloud) as guidance, we design a scene-visibility mask based latent fusion mechanism for the diffusion process of video foundation model to enhance the camera control.
By integrating user inputs, ground-adaptive 3D motion retargeting, and viewpoint-adaptive scene-visibility guided fusion, our 3STC-HIA allows flexible and adaptive control of human motion in complex 3D environments with fluent camera movement effect.

Our main contributions are summarized as follows: 
\begin{itemize}
\item We propose a highly controllable human image animation framework that integrates human action, reference image, user-defined 2D motion trajectory, and camera control signals for high-fidelity video generation;
\item We propose a ground-adaptive 3D motion retargeting approach to enable intuitive user-specified motion trajectory control by mapping plane curve to 3D world-coordinate space, adapting to the changed elevations of ground touching points of scene, and adjusting the orientation automatically;
\item We design a viewpoint-adaptive visible scene clues guided latent fusion mechanism that injects point-cloud geometric priors through scene-visibility masking, during early denoising diffusion steps, to provide precise guidance of viewpoint changes, enabling effective camera control; 
\item Extensive experimental comparisons on RealisDance-Val \cite{zhou2025realisdance-dit} and Trajectory100 \cite{liang2025realismotion} datasets show remarkable performance gains. On RealisDance-Val our approach improves key perceptual metrics, with Background Consistency 96.08 (+0.21), Motion Smoothness 98.83 (+1.22), and Aesthetic Quality 59.47 (+2.55) relative to RealisMotion \cite{liang2025realismotion}, a state-of-the-art HIA method; on Trajectory100 it achieves Translation Error 0.478 m (-0.827 m) and a Yaw Error 0.162° (+0.067°) comparable to RealisMotion.
\end{itemize}

\section{Related Work}

With the rise of diffusion models \cite{peebles2023scalable,esser2023structurecontentguidedvideosynthesis,ho2022videodiffusionmodels}, many pioneering works use 2D skeletons \cite{Yang2023Effective} as the primary pose guidance to drive motion editing and human animation. Representative methods such as \cite{musepose,zuo2024edityourmotionspacetimediffusiondecoupling,li2025tokenmotion, wang2024humanvid,ma2024followposeposeguidedtexttovideo,wang2023disco,Gait-Recognition-MIR} feed skeleton sequences into ControlNet‑style modules \cite{ZhangRA23} or pose‑guided diffusion pipelines \cite{hu2023animateanyone} to effectively guide the generation of motion video.
Due to the lack of depth information, the methods hardly handle 3D human motions and the situations with camera movements, easily producing artifacts and skeleton-appearance conflicts for complex motions under viewpoint changes \cite{tu2024motionfollowereditingvideomotion,tan2024animatexuniversalcharacterimage}.

To address these issues above,  \cite{hu2025animate2} utilizes depth maps while \cite{zhu2024champ,xu2024magicanimate,niu2025anicrafter} adopt 3D SMPL \cite{loper2015smpl} as the human representation to improve the quality of the actions in the generated videos. Realisdance-DiT \cite{zhou2025realisdance-dit}, which is built upon the video foundation model Wan-2.1 \cite{wan2025wan}, introduces additional conditional input layers and RoPE positional encoding \cite{su2021roformer}, forming a simple yet effective character animation structure. However, these methods are unable to model human body and background in a unified world coordinate space, lacking effective coupling of human and natural scene.
Uni3C \cite{cao2025uni3c} combines scene point clouds with SMPL-X \cite{pavlakos2019expressive} as joint 3D world guidance and designs a specific camera control module, but does not achieve human motion trajectory control. 
RealisMotion \cite{liang2025realismotion} emphasizes separating motion trajectories from actions in 3D world space \cite{shen2024world}, constructing a more flexible compositional and scene-aware generation pipeline. However, this approach lacks effective modeling of 3D scene information, doesn't support customized camera trajectory control, and its motion trajectory control relies on visual projection without environmental constraints. 

In summary, all these methods mentioned above are insufficient in perception of 3D structures of human and scene and their natural interactions, hardly generating photorealistic video with both human motion and camera movement \cite{cao2025uni3c}. 
Our approach comprehensively considers all these necessary elements for real human-scene-camera dynamics, achieving both controls of human motion and camera trajectories in a aligned 3D world-coordinate space and generating scene-adaptive HIA video. 

\begin{figure*}[t]
  \centering 
  \includegraphics[width=\linewidth]{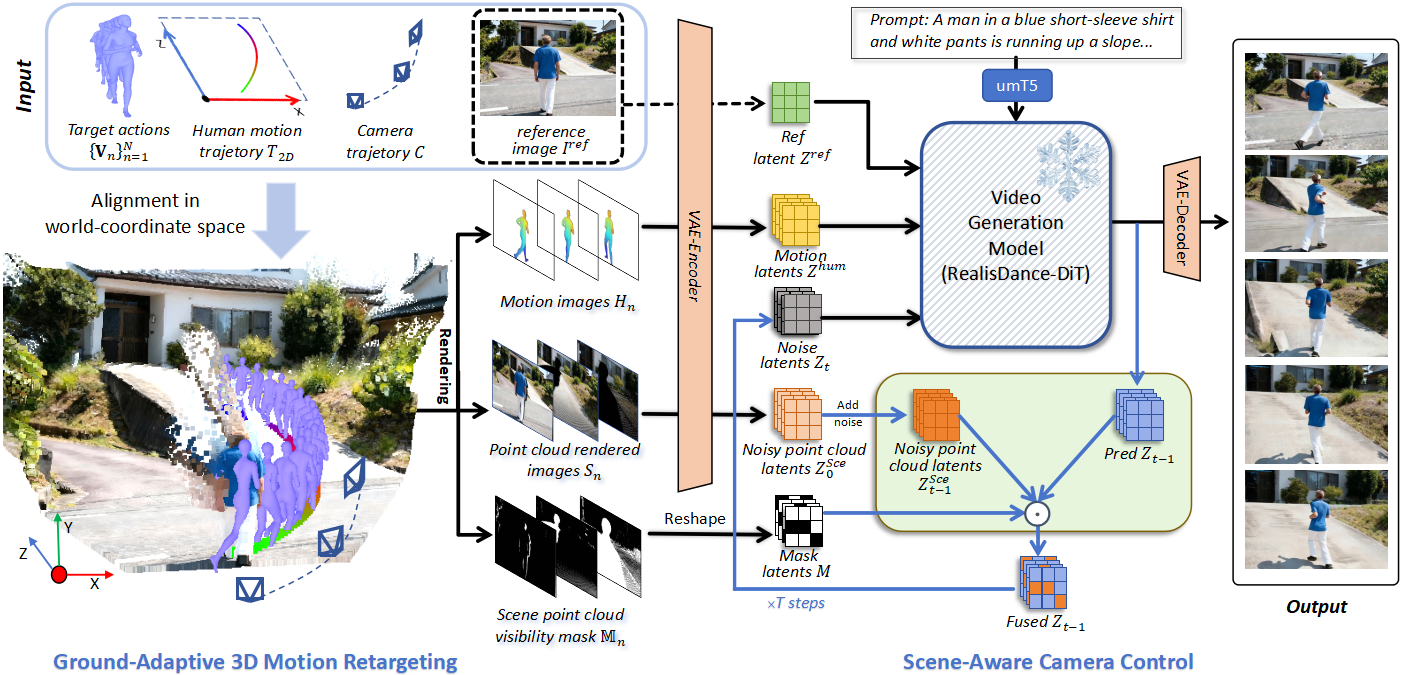} 
  \caption{\textbf{Overview of 3STC-HIA.} We first align the target action sequence $\{\mathbf{V}_n\}_{n=1}^N$ with the reconstructed point cloud of the reference image $I^{\mathrm{ref}}$ in the world-coordinate space. The subject in $I^{\mathrm{ref}}$ (e.g., the human image) is then animated to follow the user-defined 2D motion trajectory $T_{2D}$ while adapting to the varying elevations of the scene ground. During the subsequent generation stage, at each diffusion timestep, the predicted latents are fused with the noisy point cloud latents to acquire the guidance from the viewpoint-adaptive visible scene point cloud. This is achieved through a scene-visibility mask, which enhances camera control along the specified trajectory $\mathcal{C}$.} 
  \label{fig:m2} 
\end{figure*}

\section{Methodology}
\subsection{Problem Formulation}
Given a subject-scene reference image $I^{\mathrm{ref}}\in\mathbb{R}^{H\times W\times 3}$, a user-specified 2D human motion trajectory $T_{2D}$, a SMPL-X \cite{pavlakos2019expressive} action sequence $\{\mathbf{V}_n\}_{n=1}^N$, and a camera control signal $\mathcal{C}=(K,\{E_n\}_{n=1}^N)$ (parameterized in camera intrinsics $K$ and $N$ frames of extrinsic $E_n=[R_n\mid T_n]$, where $R_n$ denotes rotation and $T_n$ translation, and the frame number, i.e, the number of camera viewpoints $N$ could be different from the number of actions, but we keep the same for simplicity), we aim to synthesize a photorealistic video $\{I_n\}_{n=1}^N$ by transferring the target actions to the reference subject to make it move along the specified trajectory and adapt to scene geometry under controlled camera movement. 

To achieve this, our training-free 3STC-HIA performs a pipeline involving: 3D space representations and coordinate system alignment, 3D-to-2D projection (rendering), and pre-trained I2V diffusion model based generation guided by rendered 2D conditions. Within the aligned 3D world-coordinate space, our proposed human motion retargeting approach not only transfers the provided human actions but also enables motion trajectory control including dynamic altitude adaptation to the scene ground and orientation determination. During every diffusion step (mainly early steps), our designed camera control mechanism incorporates visible scene point cloud clues to steer the diffusion model towards viewpoint-adaptive prior-enhanced output. The overview is illustrated in \cref{fig:m2}.

\subsection{Preliminaries}
\myparagraph{3D Representation and Coordinate Alignment} To achieve accurate, flexible, multipurpose controls in complex, realistic video applications, integrating diverse inputs pre-processed by upstream models into the base model is imperative. To acquire robust 3D descriptions of human and scene with effective coordinate alignment, we adopt a series of well-studied practical pre-processing techniques. 
Specifically, to reconstruct the 3D representation of the reference image, we use Depth-Pro \cite{bochkovskii2024depth} to estimate a monocular depth map from the reference view and align it to a metric representation using SfM annotations \cite{schonberger2016pixelwise} or Multi-view Stereo reconstruction \cite{cao2024mvsformerpp}. Following \cite{cao2025uni3c}, we then generate a 3D point cloud $\mathbf{P}=\{\mathbf{p}_j\}_{j=1}^M$ of $M$ points in the world-coordinate system $\mathcal{W}$.
The target human action sequence is obtained either from a source video using~\cite{shen2024world}, or generated from textual descriptions via text-to-motion models~\cite{guo2022t2m, tevet2022mdm}. We represent each action frame using the SMPL-X model \cite{pavlakos2019expressive}, producing mesh vertices $\mathbf{V}_n \in \mathbb{R}^{V \times 3}$ in its own coordinate system $\mathcal{W_\mathrm{hum}}$.
To establish a unified coordinate space, we align the SMPL-X actions $\{\mathbf{V}_n\}_{n=1}^N$ to the reconstructed point cloud $\mathbf{P}$ in $\mathcal{W}$ by matching corresponding human body keypoints between $\mathcal{W_\mathrm{hum}}$ and $\mathcal{W}$ using a least-squares rigid transformation \cite{cao2025uni3c}. Finally, the SMPL-X gravity direction is calibrated with GeoCalib~\cite{veicht2024geocalib}.

\myparagraph{3D-to-2D Projection (Rendering)}  
To convert the aligned human action mesh $\mathbf{V}_n$ and scene point cloud $\mathbf{P}$ into 2D conditioning signals for the diffusion model, we project them into image space along the camera trajectory $\mathcal{C}=(K,\{E_n\}_{n=1}^N)$. Let $\Pi_n:\mathbb{R}^3\to\mathbb{R}^2$ denote the perspective projection operator for frame $n$, defined by the camera intrinsics $K$ and extrinsics $E_n$ corresponding to each viewpoint along $\mathcal{C}$. For every frame $n$, the rendering of each component is performed as
\begin{equation}
   H_n=\Pi_n({\textbf{V}}_n),\quad
   S_n=\Pi_n(\mathbf{P}),
\end{equation}
where $H_n$ and $S_n$ represent the projected images of the human action mesh and the scene (including the reference subject) point cloud from the $n$-th frame view respectively. The geometric condition for each frame is then defined as $\mathcal{G}_n = \{ H_n, S_n \}$, encapsulating human action, motion trajectory, camera movement, and scene geometry for conditioning diffusion process.

\myparagraph{Diffusion Backbone and Conditioning}  
We adopt the pre-trained RealisDance-DiT model~\cite{zhou2025realisdance-dit} as the backbone for video generation. 
Along with the target prompt embedding obtained from umT5~\cite{chung2023unimax}, all conditional inputs $I^{\mathrm{ref}}$, $H_n$, $S_n$ are encoded using Wan-VAE~\cite{wan2025wan}, producing $Z^{\mathrm{ref}}$, $Z^{\mathrm{hum}}_n$, $Z^{\mathrm{sce}}_n$, respectively.
The combined latent condition for each frame is then defined as $\mathbf{G}_n=\{Z^{\mathrm{hum}}_n, Z^{\mathrm{sce}}_n\}$, representing both human motion and scene geometry of the $n$-th frame. RealisDance-DiT is based on Wan2.1 \cite{wan2025wan}, which is inspired by the success of Diffusion Transformers
(DiT) \cite{peebles2023scalable} combined with Flow Matching \cite{liu2022flow}. Following WORLDFORGE \cite{song2025worldforge}, we can also view Flow Matching as a special case of diffusion modeling. Here, we take the classical DDIM sampler \cite{song2022denoising} for example to illustrate the 
sampling process. During each diffusion timestep, the latent $Z_{n,t}$ is iteratively updated conditioned on $Z^{\mathrm{ref}}$ and $\mathbf{G}_n$:

\begin{footnotesize}
\begin{equation}
   Z_{n,t-1} = \frac{1}{\sqrt{\alpha_t}} Z_{n,t} + \left( \sqrt{1-\bar{\alpha}_{t-1}} - \frac{\sqrt{1-\bar{\alpha}_t}}{\sqrt{\alpha_t}} \right) \epsilon_\theta(Z_{n,t}, t, Z^{\mathrm{ref}}, \mathbf{G}_n) ,
\end{equation}
\end{footnotesize}
where $t$ is the diffusion timestep, $\alpha_t$ denotes the noise schedule parameter and $\epsilon_\theta(\cdot)$ is the denoising network.
At the final step, the denoised latent representations are decoded to generate the output target video.




\begin{figure}[t]
  \centering

   \includegraphics[width=0.9\linewidth]{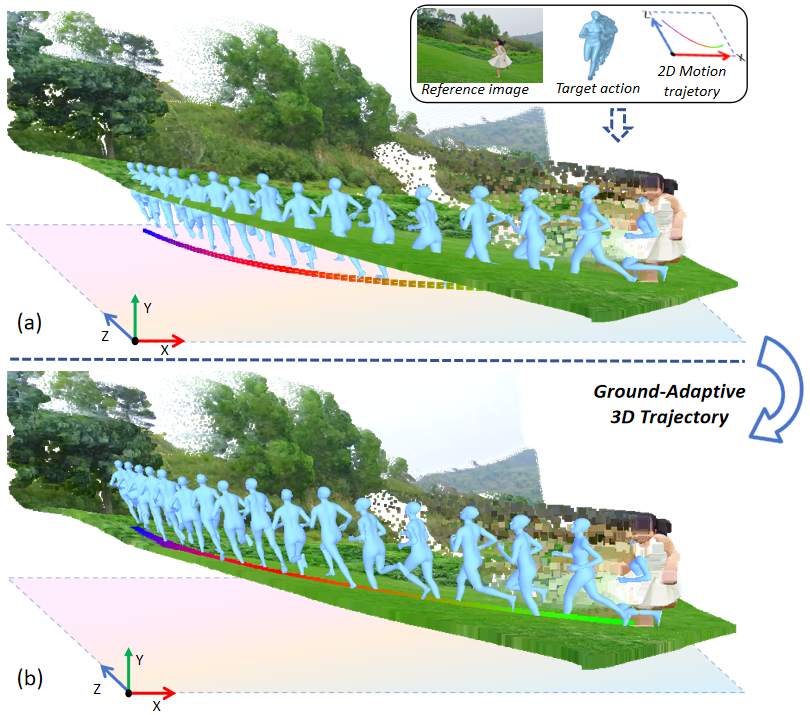}
   \caption{Human Motion Retargeting with Ground-Adaptive 3D Trajectory: (a) When transferring human actions onto scenes with uneven terrain using only a 2D motion trajectory, artifacts such as floating, ground penetration, or sinking may occur, as illustrated by the running example on a slope.
(b) In contrast, our ground-adaptive 3D trajectory control effectively adjusts the character’s motion to align with the scene’s ground elevation, producing more natural and physically consistent human animation.}
   \label{fig:height}
\end{figure}

\subsection{Ground-Adaptive 3D Motion Retargeting}
With the human action mesh and point cloud aligned in the world-coordinate space $\mathcal{W}$, beyond transferring target actions to the reference subject by adjusting the SMPL-X parameters \cite{pavlakos2019expressive}, we propose a ground-adaptive 3D motion retargeting approach to control motion displacement, altitude, and orientation in accordance with scene geometry.

\myparagraph{X-Z Displacement} As shown in \cref{fig:height}(a), given a user-defined 2D trajectory $T_{2D}$ (\eg, we can draw a curve directly on the X-Z plane of $\mathcal{W}$, i.e., the 2D top-view map of the scene), we uniformly sample $N$ points and align them one by one with the pelvis keypoints of the action frames. The resulting coordinates $\{(x_n,z_n)\}_{n=1}^N$ define the X-Z displacements of the pelvis keypoints across all frames.

\myparagraph{Adaptive Altitude} To ensure the human motion adapts to ground elevation and avoids floating or penetration, we compute the Y-axis coordinate for pelvis keypoint of each action frame based on the local ground altitude. Specifically, we estimate the natural ground height $y_n^{\mathrm{g}}$ of the ground touching keypoint in the $n-$th frame, and then based on it, adjust the Y-axis coordinate value $y_n$ of the pelvis keypoint in each aligned SMPL-X frame adaptively. 
To estimate $y_n^{\mathrm{g}}$, inspired by the bottom-few-points heuristics commonly used in LiDAR-based terrain mapping and robotic localization \cite{huber1981robust,asprs2014posacc,wilcox2017modern}, we choose to average the $Y$-axis coordinates of the points that rank the lowest $5\%$ in terms of altitude within a cylindrical neighborhood $\mathbf{P}_{\rho}$ of radius $\rho$ from the point cloud $\mathbf{P}$:
\begin{equation}
  \mathbf{P}_{\rho}= \{ \mathbf{p}_j \in \mathbf{P} \mid \|(\mathbf{p}_j)_{xz} - (x_n, z_n)\| \le \rho \,\},
\end{equation}
where, $(\mathbf{p}_j)_{xz}$ denotes the X-Z coordinate of point $\mathbf{p}_j$, and $\rho$ is set to $0.15\,\mathrm{m}$ in our experiments.
Let $\{\mathbf{r}_n^{\mathrm{ori}}\}_{n=1}^N$ be the originally SMPL-X root (pelvis) trajectory in the $\mathcal{W}$ space, $\mathbf{r}_n^{\mathrm{ori}}=(x_{n}^{\mathrm{ori}},y_{n}^{\mathrm{ori}},z_{n}^{\mathrm{ori}})$. By adding the ground altitude gap $y_n^{\mathrm{g}}-y_1^{\mathrm{g}}$ of the touching point in the $n$-th frame compared to the start point in the first frame, we update the 3D root trajectory into $\{\mathbf{r}_n\}_{n=1}^N$:
\begin{equation}
\mathbf{r}_n = (x_n,\; y_{n}^{\mathrm{ori}} + (y_n^{\mathrm{g}}-y_1^{\mathrm{g}}),\; z_n),
\end{equation}
which denotes the new pelvis keypoint coordinate, navigating the human motion adaptively to the uneven ground while preserving the source motion's height dynamics, as shown in \cref{fig:height}(b).

\myparagraph{Orientation}
For the \(n\)-th frame, the forward tangent vector of the updated 3D human motion trajectory is computed as
\begin{equation}
\mathbf{t}_n = \frac{\mathbf{r}_{n+1} - \mathbf{r}_n}{\|\mathbf{r}_{n+1} - \mathbf{r}_n\|}.
\end{equation}
The corresponding yaw angle is obtained by
\begin{equation}
\theta_n = \mathrm{atan2}(t_{n,x},\, t_{n,z}),
\end{equation}
which measures the rotation around the Y-axis. To ensure smooth orientation transition, the sequence \(\{\theta_n\}\) is then temporally smoothed with a Gaussian kernel (\(\sigma=3\)), and the resulting smoothed yaw angle \(\tilde{\theta}_n\) defines a rotation matrix around the vertical Y-axis:
\begin{equation}
\mathbf{R}_y(\tilde{\theta}_n) =
\begin{bmatrix}
\cos\tilde{\theta}_n & 0 & \sin\tilde{\theta}_n \\
0 & 1 & 0 \\
-\sin\tilde{\theta}_n & 0 & \cos\tilde{\theta}_n,
\end{bmatrix}
\end{equation}
which controls the final orientation of the 3D human motion trajectory.

\begin{figure}[t]
  \centering
   \includegraphics[width=0.9\linewidth]{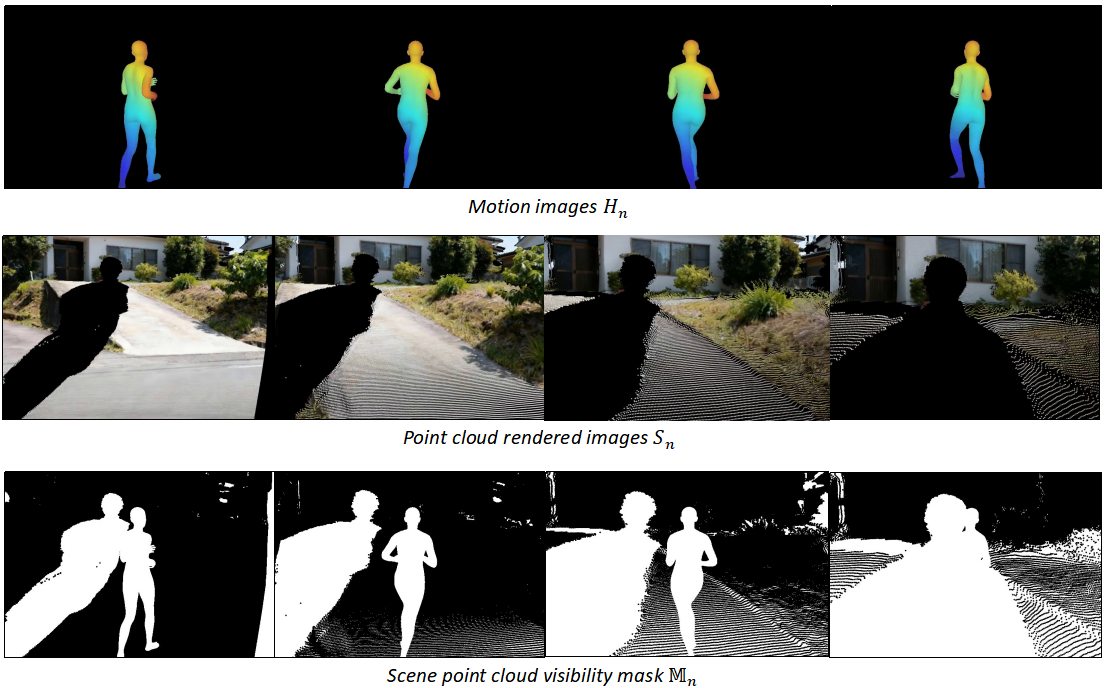}

   \caption{Illustration of viewpoint-adaptive scene point cloud visibility mask. }
   \label{fig:mask}
\end{figure}

\subsection{Scene-Aware Camera Control}

In addition to controlling human motion trajectory, we also aim to produce realistic camera movement in the generated video. Along the user-specified camera trajectory, sequences of rendered images $\{ H_n \}_{n=1}^N$ and $\{ S_n \}_{n=1}^N$ are obtained via 3D-to-2D projection. Instead of simply concatenating or overlaying $\{ H_n \}_{n=1}^N$ and $\{ S_n \}_{n=1}^N$ as conditional guidance, we design a scene-aware camera control mechanism that enhances realism by fusing viewpoint-adaptive visible point cloud information through a visibility mask–based latent fusion process.

\myparagraph{Scene Visibility Mask Computation}  
For the $n$-th viewpoint, we project both the aligned 3D human motion and scene point cloud using the same camera parameters $\Pi_n$, resulting in rendered images $H_n$ and $S_n$. At each image coordinate $(u,v)$, we compute two depth values, $d_H(u,v)$ and $d_S(u,v)$, representing the distances from the camera center to the nearest visible human and scene surfaces, respectively ($d_*(u,v)=\infty$ when no visible point exists). As shown in \cref{fig:mask}, a viewpoint-adaptive binary mask $\mathbb{M}_n(u,v)$ is generated as
\begin{equation}
 \mathbb{M}_n(u,v)=
\begin{cases}
1, & \text{if } d_S(u,v) < d_H(u,v) \text{ and } d_S(u,v) < \infty \\
0, & \text{otherwise}
\end{cases}   
\end{equation}
which distinguishes visible scene regions from occluded or human-occupied areas in the monocular view.
\begin{figure}[t]
  \centering 
  \includegraphics[width=\linewidth]{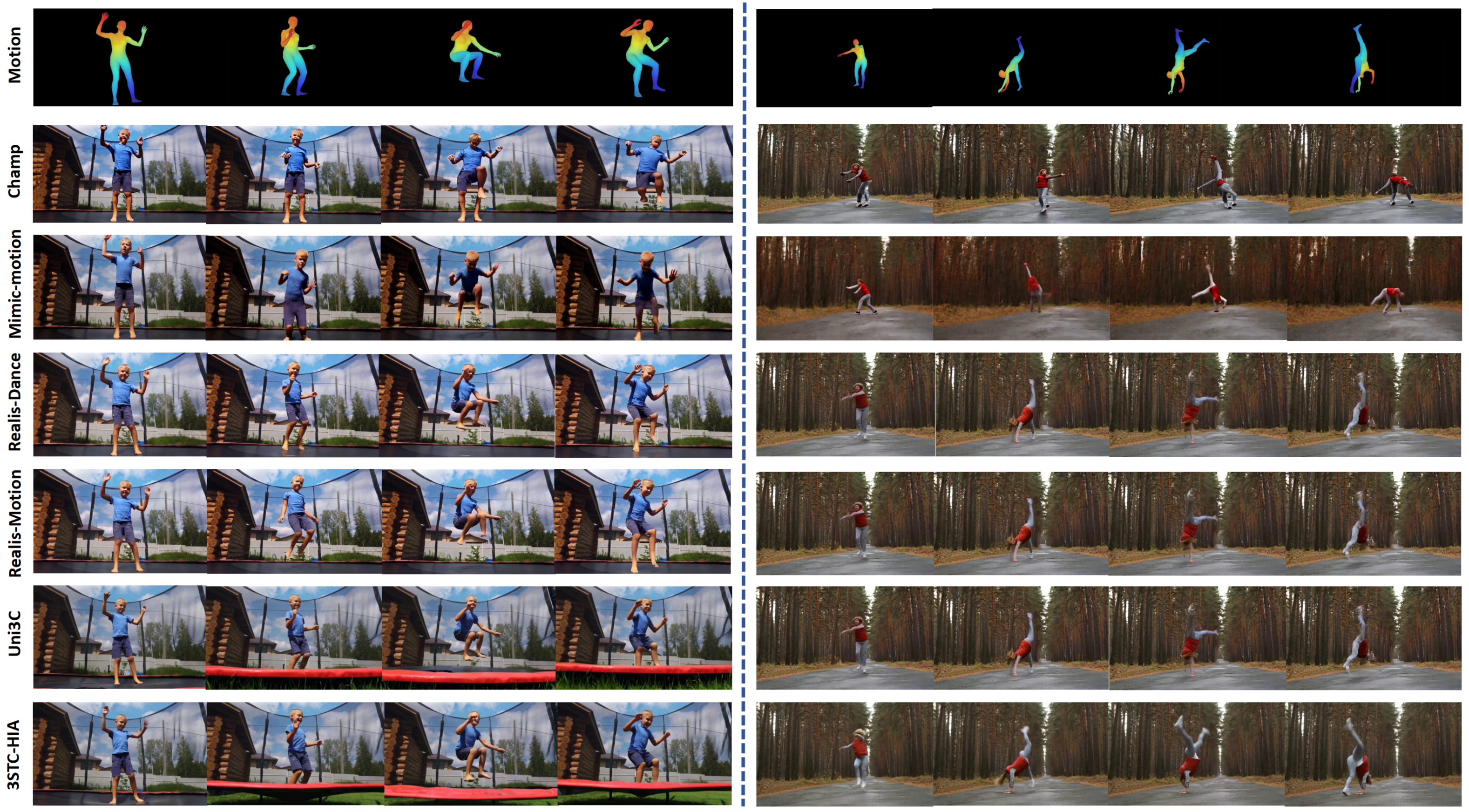} 
  \caption{Qualitative comparison of our method and baselines. The results of our method demonstrate coherent and natural human motion that maintains good physical consistency with the scene.} 
  \label{fig:e1} 
\end{figure}

\myparagraph{Viewpoint-Adaptive Scene Guided Fusion} To generate more photorealistic video with specified camera movement, we inject viewpoint-adaptive real scene information into the diffusion process. Specifically, we fuse the predicted latents with the latents derived from the visible scene point cloud using the correspondingly reshaped mask $M_n$, in every denoising timestep for every frame, continuously enhancing the real scene adaptivity. 
In each timestep, to match with the predicted latents $Z_{t-1}$ to achieve effective fusion, we first forward-diffuse the Wan-VAE encoded point cloud latents $Z_0^{\mathrm{sce}}$, adding noise to the same level with $Z_{t-1}$ (for simplicity, we omit the subscript $n$ indicating the frame number):
\begin{equation}
Z_{t}^{\mathrm{sce}} = \sqrt{\bar\alpha_t}\,Z_0^{\mathrm{sce}} + \sqrt{1-\bar\alpha_t}\,\epsilon,\qquad \epsilon\sim\mathcal{N}(0,I)
\end{equation}


To implement a strong-to-weak scene visibility control, we also introduce a temporal decaying weight $\lambda(t)$, so the mask based fusion can be expressed as: 
\begin{equation}
    Z_{t} \leftarrow \bigl(1 - \lambda(t) M\bigr) \odot Z_{t} \;+\; \lambda(t) M \odot Z_{t}^{\mathrm{sce}}
\end{equation}
where $\odot$ denotes element-wise multiplication. 
We employ an exponential decay: $\lambda(t) = \lambda_{\mathrm{max}} \exp(\beta(t-T))$, where $\lambda_{\mathrm{max}}\in[0.7,0.95]$, and $\beta>0$ is a hyperparameter controlling decay speed (typically $\beta \approx 0.1\sim 0.3$ for $T=50$ steps).

In this fusion mechanism, the rendered scene latent $Z_t^{\mathrm{sce}}$ inherently encodes the truly visible scene point cloud clues from the changing viewpoints along the specified camera trajectory, enhancing the guidance of the geometric priors of scene\footnote{Actually, the viewpoint-adaptive visible scene point cloud also involves partial 3D points restructured from the subject of reference image, while their impact on the generated effectiveness can be overlooked as verified in \cite{cao2025uni3c}.}. The denoised $Z_t$ contains the guided information of reference subject's appearance and the human motion. The mask $M$ achieves effective information fusion while highlighting their respective roles.
The decaying $\lambda(t)$ over time, considers to make the diffusion generation in early time steps strongly adhere to the visible geometric priors to stabilize global scene structure, while allowing for free synthesis of natural human appearance and dynamics with progressively enhanced fine-grained details in late time steps.

\section{Experiments}

\begin{table}[t]
  \centering 
  \resizebox{\textwidth}{!}{
    \footnotesize   
    \setlength{\tabcolsep}{2pt}  
    \begin{tabular}{@{}l c c c *{6}{c} c c@{}}
      \toprule 
      Method & \thead{Action \\ Control} & \thead{Motion \\ Trajectory} & \thead{Camera \\ Trajectory} & \thead{Subject \\ Consistency$\uparrow$} & \thead{Background \\ Consistency$\uparrow$} & \thead{Temporal \\ Flicker$\uparrow$} & \thead{Motion \\ Smoothness$\uparrow$} & \thead{Aesthetic \\ Quality$\uparrow$} & \thead{Dynamic \\ Degree$\uparrow$} & \thead{FID$\downarrow$} & \thead{FVD$\downarrow$} \\ 
      \midrule 
      ControlNeXt \cite{peng2024controlnext} & \ding{51} & \ding{55} & \ding{55} & 92.15 & 94.21 & 96.45 & 97.83 & 54.89 & 61.37 & 36.47 & 798.15 \\ 
      MusePose \cite{musepose} & \ding{51} & \ding{55} & \ding{55} & 93.42 & 95.16 & 97.23 & 98.31 & 55.76 & 58.94 & 41.28 & 1035.71 \\ 
      Champ \cite{zhu2024champ} & \ding{51} & \ding{55} & \ding{55} & 92.87 & 94.05 & 95.64 & 97.12 & 55.43 & 67.15 & 39.84 & 856.92 \\ 
      MimicMotion \cite{zhang2025mimicmotionhighqualityhumanmotion} & \ding{51} & \ding{55} & \ding{55} & 91.75 & 93.84 & 96.52 & 98.07 & 53.86 & 60.42 & 34.79 & 775.68 \\ 
      RealisDance \cite{zhou2025realisdance-dit} & \ding{51} & \ding{55} & \ding{55} & 94.26 & 95.97 & 97.68 & 98.59 & 57.35 & 65.28 & 25.13 & \textbf{305.84} \\ 
      HumanVid \cite{wang2024humanvid} & \ding{51} & \ding{55} & \ding{51} & 92.33 & 93.66 & 95.37 & 98.64 & 53.78 & 61.16 & 36.84 & 877.52 \\ 
      RealisMotion \cite{liang2025realismotion} & \ding{51} & \ding{51} & \ding{55} & \textbf{94.58} & 95.87 & 96.12 & 97.61 & 56.92 & 64.17 & 24.10 & 519.66 \\ 
      Uni3C \cite{cao2025uni3c} & \ding{51} & \ding{55} & \ding{51} & 92.68 & 93.47 & \textbf{97.82} & 97.95 & 57.61 & \textbf{73.24} & \textbf{23.86} & 516.36 \\ 
      \midrule 
       \textbf{3STC-HIA} & \ding{51} & \ding{51} & \ding{51} & 94.13 & \textbf{96.08} & 97.51 & \textbf{98.83} & \textbf{59.47} & 72.36 & 27.92 & 573.41 \\ 
      \bottomrule 
    \end{tabular}%
  }
  \caption{Evaluation on RealisDance-Val\cite{zhou2025realisdance-dit} using the Vbench-I2V \cite{vbench2024} perceptual dimensions (6 middle columns) and FID/FVD (last 2 columns). For the perceptual columns, the higher the value, the better. For FID/FVD, the lower, the better.} 
  \label{tab:realisdance_vbench_en}
\end{table}

We evaluate our approach by comparing with state-of-the-art methods in terms of generation quality and trajectory controllability metrics. We also conduct ablative studies on two core components as well as different subsets of control signals to evaluate their individual contributions. All methods use the relatively equivalent inputs as needed respectively and identical evaluation metrics. More details and visual results can be found in supplementary materials.
\subsection{Experimental Settings}
\myparagraph{Datasets}
We evaluate on two benchmark datasets. RealisDance-Val~\cite{zhou2025realisdance-dit}: a 100-clip validation set collected from Internet videos, covering diverse scenes, characters, poses, lighting, and interactions. For each clip, we use the first frame as the reference image. Trajectory100~\cite{liang2025realismotion}: a 100-clip benchmark to evaluate global trajectory and orientation control. For each clip, we reconstruct the ground-truth 3D motion trajectory and yaw sequence as control signals, and use the first frame as reference. 

\myparagraph{Baselines}
We compare 3STC-HIA with other human image animation methods, including ControlNeXt~\cite{peng2024controlnext}, MusePose~\cite{musepose}, Champ~\cite{zhu2024champ}, MimicMotion~\cite{zhang2025mimicmotionhighqualityhumanmotion}, HumanVid \cite{wang2024humanvid}, Uni3C~\cite{cao2025uni3c}, RealisDance-DiT~\cite{zhou2025realisdance-dit}, and RealisMotion~\cite{liang2025realismotion} on generation quality. Since there are relatively few methods specifically designed for controlling human motion trajectories, we additionally compare motion trajectory control capability with Tora~\cite{zhang2025tora} (trajectory-focused generation), and with RealisDance~\cite{zhou2025realisdance-dit} and RealisMotion~\cite{liang2025realismotion} from the human animation literature.


\myparagraph{Metrics} For  \textbf{generation quality}: we measure 6 perceptual dimensions (Subject Consistency, Background Consistency, Temporal Flicker, Motion Smoothness, Aesthetic Quality, Dynamic Degree) based on Vbench-I2V~\cite{vbench2024}, as well as FID \cite{heusel2018ganstrained} and FVD\cite{unterthiner2019accurate}. For \textbf{motion trajectory control}: following RealisMotion \cite{liang2025realismotion}, we extract 3D trajectories from the generated videos and align them with the ground-truth trajectories to compute the translation and rotation errors defined by MotionCtrl \cite{wang2024motionctrl}: TransError (m) is the average per-frame Euclidean distance, and RotError (deg) is the average absolute yaw difference.


\begin{figure}[t] 
  \centering 
   \includegraphics[width=1\linewidth]{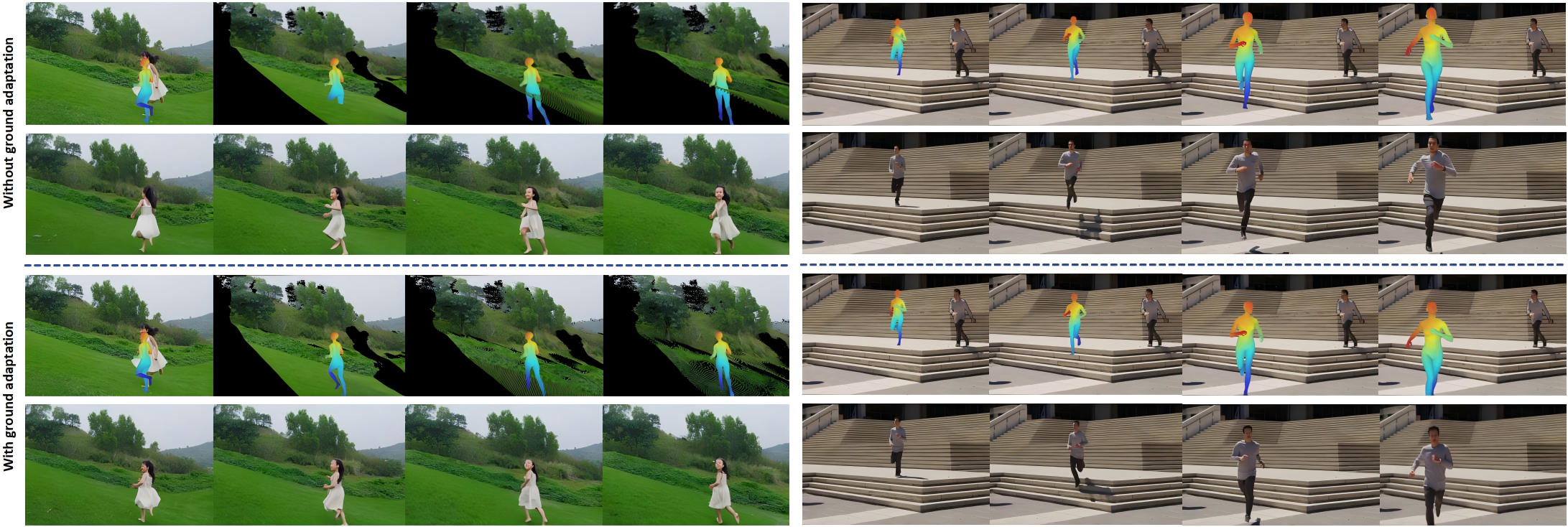} 
   \caption{Ablation on ground-adaptive 3D motion retargeting. Our method addresses floating or penetrating on slopes or steps while enabling the human movement to undulate with the ground.}
   \label{fig:abl1} 
\end{figure} 

\begin{figure}[t] 
  \centering 
   \includegraphics[width=1\linewidth]{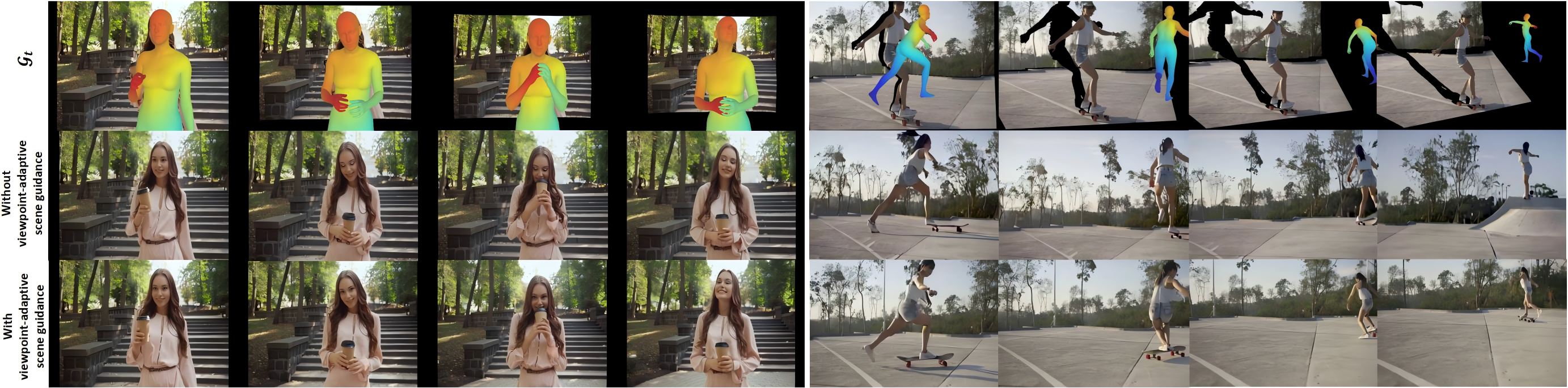} 
   \caption{Ablation on viewpoint-adaptive scene guided fusion for camera control. \textbf{Left}: When the camera pulls back to follow a forward-moving character, our method preserves correct background recession and perspective, avoiding first-frame dominance. \textbf{Right}: Even under significant camera movement, it maintains smooth motion and physically consistent background.} 
   \label{fig:abl3} 
\end{figure} 
\subsection{Comparison with state-of-the-art methods} 
\Cref{tab:realisdance_vbench_en} presents the evaluation results on the RealisDance-Val dataset. Our method achieves optimal or near-optimal performances across 6 perceptual metrics.
The proposed scene-visibility guided fusion mechanism stabilizes scene geometry under changed camera viewpoints, effectively eliminating background drift and flicker. Meanwhile, the ground-adaptive trajectory control preserves natural actions and improves human-scene coherence, yielding more physically realistic motion videos. Although our FID/FVD scores are slightly worse than some baselines, it remains reliably competitive. It is comprehensible
since our model prioritizes accurate controls of motion trajectory and scene layout fidelity over matching of statistical distributions (generated vs real data) which FID/FVD highlights.

\Cref{tab:traj} reports results on the Trajectory100 dataset, where our method significantly reduces translation drift while maintaining competitive yaw accuracy, demonstrating faithful control of motion trajectories. A qualitative comparison in \cref{fig:e1} further illustrates these advantages, showing more photorealistic human motion and stronger human-scene coherence than the baselines.
\vspace{-8pt}
\begin{table}[t] 
\centering 
\small 
\setlength{\tabcolsep}{4pt} 
\begin{tabular}{lcc} 
\toprule 
Method & Trans.\ Error (m)$\downarrow$ & Rot.\ Error (deg)$\downarrow$ \\ 
\midrule 
Tora \cite{zhang2025tora}& 5.923 & 0.342 \\ 
RealisDance-DiT \cite{zhou2025realisdance-dit}& 1.815 & 0.158 \\ 
RealisMotion \cite{liang2025realismotion} & 1.305 & \textbf{0.095} \\ 
\midrule
 \textbf{3STC-HIA} & \textbf{0.478} & 0.162\\ 
\bottomrule 
\end{tabular}  
\caption{Motion trajectory control performance comparisons on Trajectory100.}
\label{tab:traj}
\end{table}
\vspace{-8pt}
\subsection{Ablation Studies} 

We conduct comprehensive ablative experiments on: (1) the proposed ground-adaptive 3D motion retargeting, (2) viewpoint-adaptive scene-guided fusion for camera control, and (3) the decoupling and combinability of different control signal subsets.

\myparagraph{Ground-Adaptive 3D Motion Retargeting} As shown in \cref{fig:abl1}, without ground-adaptive 3D motion retargeting, the transferred actions often float above or penetrate into the inclined surfaces of scenes such as slopes or steps. In contrast, our retargeting effectively eliminates vertical drift and enforces stronger physical consistency with the scene geometry, while preserving the original action dynamics, resulting in more natural and coherent motion videos.

\myparagraph{Scene Guided Fusion for Camera Control} 
Image-conditioned video diffusion models often suffer from first-frame dominance, overly relying on the conditioning image and suppressing appearance or viewpoint changes \cite{zhao2024leakage, tian2025extrapolating, wu2023freeinit}. This is especially problematic when substantial camera movement is required. Our method introduces point-cloud geometric priors guided scene-visibility masking mechanism to preserve the intended camera movements, resulting in smooth videos even under large viewpoint changes (\cref{fig:abl3}).

\Cref{tab:ablate} reports the ablation results on RealisDance-Val. Removing the ground-adaptive 3D motion retargeting leads to increased vertical drift and slight degradation in motion smoothness and dynamics, while omitting the viewpoint-adaptive scene-guided fusion remarkably reduces the background consistency and FID/FVD performance. Overall, the complete model benefits from the complementary strengths of both components, achieving more realistic motion and improved video generation controllability.
\begin{table}[t]
  \centering
  \resizebox{\textwidth}{!}{%
    \small
    \setlength{\tabcolsep}{2.5pt}
    \begin{tabular}{@{}lcccccccc@{}}
      \toprule
       & \begin{tabular}[c]{@{}c@{}}Subject\\Consist.$\uparrow$\end{tabular}
       & \begin{tabular}[c]{@{}c@{}}Background\\Consist.$\uparrow$\end{tabular}
       & \begin{tabular}[c]{@{}c@{}}Temporal\\Flicker$\uparrow$\end{tabular}
       & \begin{tabular}[c]{@{}c@{}}Motion\\Smooth.$\uparrow$\end{tabular}
       & \begin{tabular}[c]{@{}c@{}}Aesthetic\\Quality$\uparrow$\end{tabular}
       & \begin{tabular}[c]{@{}c@{}}Dynamic\\Degree$\uparrow$\end{tabular}
       & FID$\downarrow$
       & FVD$\downarrow$ \\
      \midrule
      \makecell[l]{w/o 3D ground\\adaptive retargeting} 
        & 91.35 & 94.82 & 95.47 & 93.28 & 56.42 & 65.73 & 41.65 & 685.42 \\
      \makecell[l]{w/o viewpoint\\adaptive guidance} 
        & 91.74 & 90.23 & 90.56 & 96.35 & 53.28 & 64.82 & 58.47 & 832.15 \\
      \midrule
      Full model & 93.12 & 95.64 & 96.89 & 97.42 & 57.86 & 68.95 & 32.78 & 521.37 \\
      \bottomrule
    \end{tabular}
  }
  \caption{Ablation study on RealisDance-Val. We analyze the roles of the two key components: 3D ground-adaptive retargeting and view-adaptive scene guidance.}
  \label{tab:ablate}
\end{table}

\begin{figure}[t] 
  \centering 
   \includegraphics[width=1\linewidth]{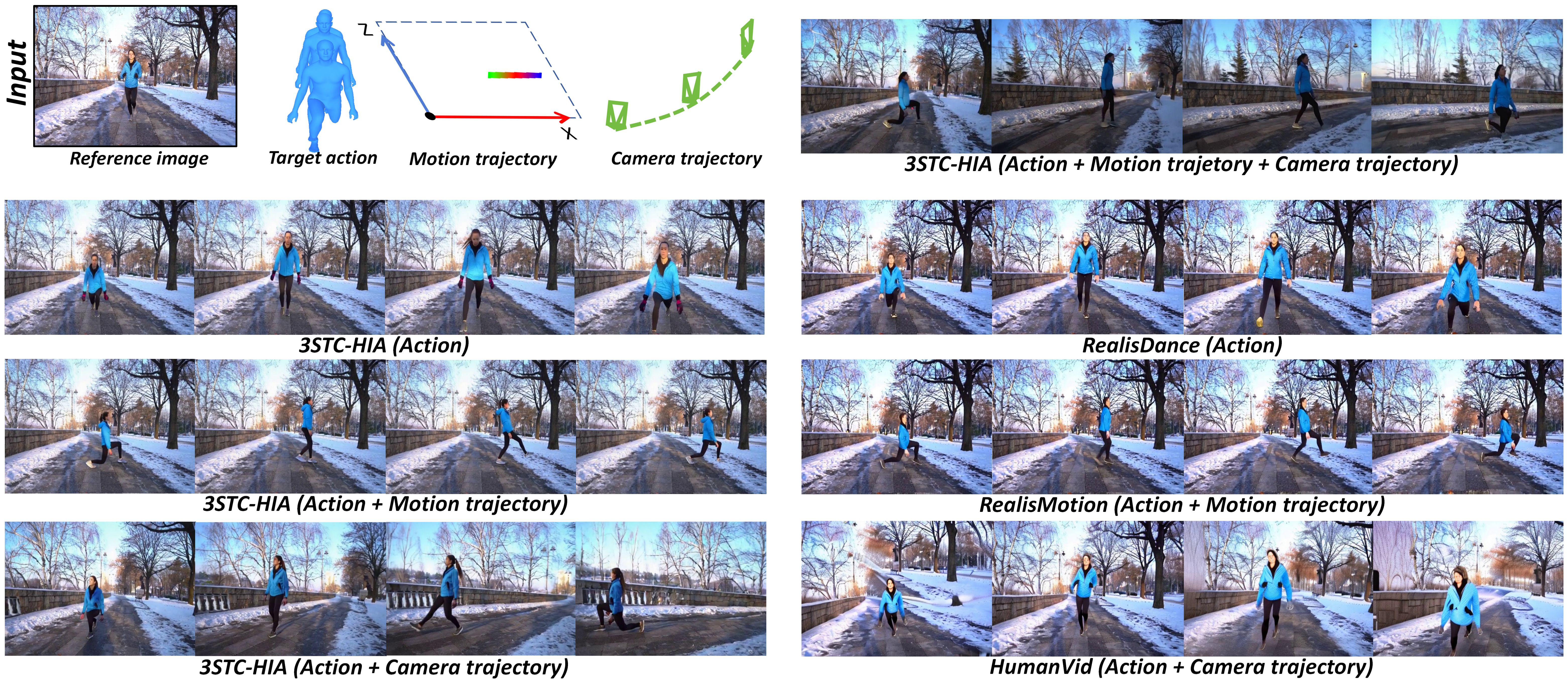} 
   \caption{Ablation of Different Control Combinations. Different control signals can be flexibly combined to meet the complex requirements of practical applications.}
   \label{fig:abc} 
\end{figure}

\myparagraph{Ablation of Different Control Combinations} 
\cref{fig:abc} presents the ablation experiments of different subsets of control signals and comparisons with existing counterpart methods, verifying the effectiveness of each independent control signal of our method. The experiments show that different control signals could be decoupled, and can be flexibly combined to meet the control requirements of complex real scenarios. Compared with existing approaches, 3STC-HIA does not need to train/fine-tune independent modules for different control tasks, and can simultaneously achieve fine-grained action driving, precise motion trajectory planning, and flexible camera tracking in a same scene, without interference between the signals. This decoupling and combinability features provide greater creative freedom and more precise visual controllability for practical applications such as film preview, virtual production, and interactive content generation.

\section{Conclusion}
In this work, we present a novel framework for HIA, achieving controls of both motion trajectory and camera dynamics within complex 3D environments. A ground-adaptive 3D motion retargeting approach is developed to enable user-defined motion trajectory control, adapting to the changed elevations of ground touching points of scene and adjusting the orientation automatically. Derived from the viewpoint-adaptive visible scene clues, a novel latent fusion mechanism is designed to inject point-cloud geometric priors through scene-visibility masking, enabling effective camera control. In future, we will enhance the human-scene interactions within complex environment and consider multi-person coordinated motion control for the HIA task.

\myparagraph{Acknowledgements} Thank the support of China Scholarship Council (CSC) for providing research fellowship for the first author. This work was also supported in part by the National Natural Science Foundation of China under Grant 62372150.
This research was also supported in part by the UKRI-AHRC CoSTAR National Lab for Creative Industries Research and Development (AH/ Y001060/1).


%
%
\bibliographystyle{splncs04}
\bibliography{main}

\end{document}